\documentclass[letterpaper, 10 pt, conference]{ieeeconf}  %

\IEEEoverridecommandlockouts                              %

\usepackage{mathtools}
\usepackage{graphicx}
\graphicspath{{img/}}
\usepackage[svgpath=img]{svg}
\usepackage{epstopdf}
\usepackage{pgf}
\usepackage{calc}
\newcommand\inputpgf[2]{{
\let\pgfimageWithoutPath\pgfimage
\renewcommand{\pgfimage}[2][]{\pgfimageWithoutPath[##1]{#1/##2}}
\input{#1/#2}
}}

\usepackage{todonotes}
\usepackage{soul}
\usepackage{etoolbox}
\usepackage{amsmath}
\usepackage{amssymb}
\usepackage{bm}
\usepackage[export]{adjustbox}
\usepackage{tabularx}
\newcolumntype{Y}{>{\centering\arraybackslash}X} %
\usepackage{tikz}
\usetikzlibrary{bayesnet}
\newlength{\state} %
\usepackage{url}
\usepackage{array}
\usepackage{textcomp}
\usepackage{xspace}
\usepackage{booktabs}
\usepackage{tablefootnote}
\usepackage{cite}
\usepackage{multirow}
\usepackage{makecell}
\let\labelindent\relax
\usepackage{enumitem}
\setlist[itemize]{leftmargin=*}
\usepackage{comment}
\usepackage{color}

\newcommand{\blue}[1]{{\color{blue}#1}}
\newcommand{\green}[1]{{\color{green}#1}}
\newcommand{\red}[1]{{\color{red}#1}}

\newcommand{\LC}[1]{\red{{\bf LC:}~#1}}

\usepackage{hyperref}
\hypersetup{
    colorlinks,
    linkcolor={red!50!black},
    citecolor={blue!50!black},
    urlcolor={blue!80!black}
}
\usepackage[capitalize]{cleveref}
\usepackage{xspace}
\renewcommand{\Cref}[1]{\cref{#1}} %
\newcommand{\specialcell}[2][c]{%
  \begin{tabular}[#1]{@{}c@{}}#2\end{tabular}}
 
\usepackage[font=small,labelfont=bf]{caption}
\usepackage{pifont}%
\newcommand{\cmark}{\ding{51}}%
\newcommand{\xmark}{\ding{55}}%

\newcommand{\Kimera}{{Kimera}\xspace}
\newcommand{\KimeraVIO}{{Kimera-VIO}\xspace}
\newcommand{\KimeraRPGO}{{Kimera-RPGO}\xspace}
\newcommand{\KimeraMesher}{{Kimera-Mesher}\xspace}
\newcommand{\KimeraSemantics}{{Kimera-Semantics}\xspace}

\newcommand{\perFrameMeshLatency}{20\text{ms}}
\newcommand{\globalMeshLatency}{0.1\text{s}}

\newcommand{\KimeraUrl}{{\small\href{https://github.com/MIT-SPARK/Kimera}{https://github.com/MIT-SPARK/Kimera}}}
\newcommand{\VideoUrl}{{\small\href{https://www.youtube.com/watch?v=-5XxXRABXJs}{https://www.youtube.com/watch?v=-5XxXRABXJs}}}
\newcommand{\hide}[1]{}
\newcommand{\ie}{\emph{i.e.,~}}
\newcommand{\eg}{\emph{e.g.,~}}

\newcommand{\bdmath}{\begin{dmath}}%
\newcommand{\edmath}{\end{dmath}}
\newcommand{\beq}{\begin{equation}}
\newcommand{\eeq}{\end{equation}}
\newcommand{\bdm}{\begin{displaymath}}
\newcommand{\edm}{\end{displaymath}}
\newcommand{\bea}{\begin{eqnarray}}
\newcommand{\eea}{\end{eqnarray}}
\newcommand{\beal}{\beq \begin{array}{ll}}
\newcommand{\eeal}{\end{array} \eeq}
\newcommand{\beas}{\begin{eqnarray*}}
\newcommand{\eeas}{\end{eqnarray*}}
\newcommand{\ba}{\begin{array}} %
\newcommand{\ea}{\end{array}}

\newcommand{\bit}{\begin{itemize}}
\newcommand{\eit}{\end{itemize}}
\newcommand{\ben}{\begin{enumerate}}
\newcommand{\een}{\end{enumerate}}
\newcommand{\bbmat}{\begin{bmatrix}}
\newcommand{\ebmat}{\end{bmatrix}}
\newcommand{\bpmat}{\begin{pmatrix}}
\newcommand{\epmat}{\end{pmatrix}}

\newcommand{\Real}{\mathbb{R}}

\newcommand{\SEthree}{\ensuremath{\mathrm{SE}(3)}\xspace}

\newcommand{\MA}{\bm{A}}

\newcommand{\Euroc}{EuRoC\xspace}

\newcommand{\linkToPdf}[1]{\href{#1}{\blue{(pdf)}}}
\newcommand{\linkToPpt}[1]{\href{#1}{\blue{(ppt)}}}
\newcommand{\linkToCode}[1]{\href{#1}{\blue{(code)}}}
\newcommand{\linkToWeb}[1]{\href{#1}{\blue{(web)}}}
\newcommand{\linkToVideo}[1]{\href{#1}{\blue{(video)}}}
\newcommand{\award}[1]{\xspace} %

\newcommand{\myparagraph}[1]{{\bf#1.}}

\newcommand{\etal}{\emph{et al.}\xspace}

 \newcommand{\myNo}{\red{\xmark}}
\newcommand{\myYes}{\green{\cmark}}

\newcommand{\toAdd}[1]{}
\newcommand{\isExtended}[2]{#2}

\newcommand{\Toni}[1]{#1}

\renewcommand{\st}[1]{} %

\newcommand{\myClaim}{an\xspace}

\newcommand{\vinVersion}[2]{#2} %

\newcommand{\gtwoo}{{g2o}\xspace}%

\title{\LARGE \bf
\Kimera: an Open-Source Library for Real-Time \\ Metric-Semantic  Localization and Mapping \vspace{-2mm} %
}

\author{Antoni Rosinol, Marcus Abate, Yun Chang, Luca Carlone
\thanks{A.\,Rosinol, M.\,Abate, Y.\,Chang, L.\,Carlone are with the Laboratory for
Information \& Decision Systems (LIDS), Massachusetts Institute of Technology, Cambridge, MA, USA,
{\sf\scriptsize \{arosinol,mabate,yunchang,lcarlone\}@mit.edu}}
\thanks{This work was partially funded by ARL DCIST CRA W911NF-17-2-0181, MIT Lincoln Laboratory, and `la Caixa' Foundation (ID 100010434), under the agreement LCF/BQ/AA18/11680088 (A. Rosinol).}}

\begin{document}

\makeatletter
\let\@oldmaketitle\@maketitle%
\renewcommand{\@maketitle}{\@oldmaketitle%

  \begin{minipage}{\textwidth}
  \begin{center}
  \begin{tabular}{cc}
  \hspace{-2mm}\includegraphics[trim={0cm 0cm 0cm 0cm}, clip, width=0.99\columnwidth]{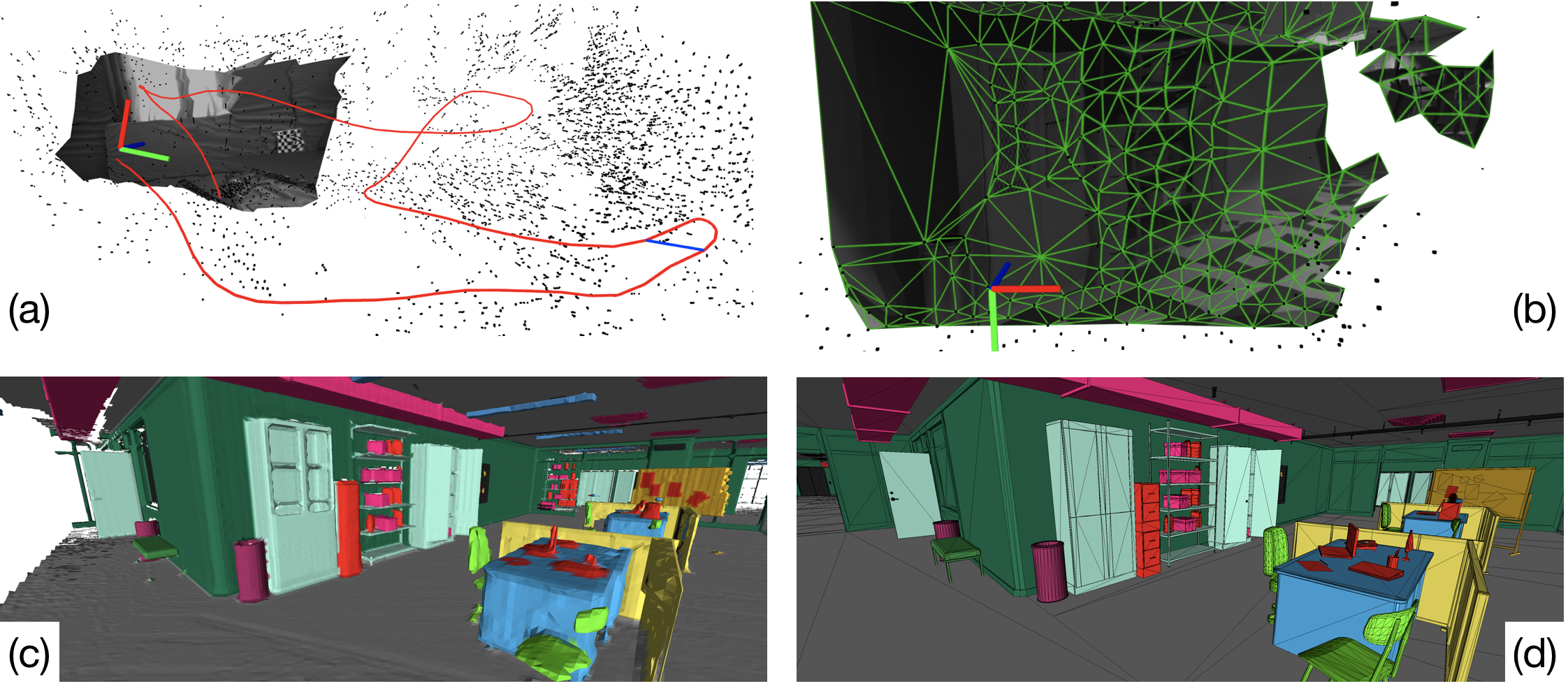}
  \end{tabular}
   \end{center}
   \vspace{-2mm}
  \end{minipage}
\\
{\small{\bf Fig.\,1:} \Kimera is an open-source C++ library for real-time metric-semantic SLAM. It provides
 (a) visual-inertial state estimates at IMU rate, and
 a globally consistent and outlier-robust trajectory estimate, 
 computes (b) a low-latency local mesh of the scene that can be used for fast obstacle avoidance, 
 and  builds (c) a global semantically annotated 3D mesh, which accurately reflects\Toni{\st{ (d)}} the ground truth model \Toni{(d)}. 
 \vspace{-7mm}} 
  }%
\makeatother

\maketitle

\begin{tikzpicture}[overlay, remember picture]
\path (current page.north east) ++(-6.3,-0.0) node[below left] {
Accepted for publication at ICRA 2020, please cite as follows:
};
\end{tikzpicture}
\begin{tikzpicture}[overlay, remember picture]
\path (current page.north east) ++(-7.2,-0.4) node[below left] {
A. Rosinol, M. Abate, Y. Chang, L. Carlone
};
\end{tikzpicture}
\begin{tikzpicture}[overlay, remember picture]
\path (current page.north east) ++(-4.6,-0.8) node[below left] {
``Kimera: an Open-Source Library for Real-Time Metric-Semantic Localization and Mapping'',
};
\end{tikzpicture}
\begin{tikzpicture}[overlay, remember picture]
\path (current page.north east) ++(-7.5,-1.2) node[below left] {
 IEEE Int. Conf. Robot. Autom. (ICRA), 2020.
};
\end{tikzpicture}

\setcounter{figure}{1}

\begin{abstract}
We provide \myClaim open-source C++ library for real-time
 metric-semantic visual-inertial Simultaneous Localization And Mapping (SLAM). 
The library goes beyond existing visual and visual-inertial SLAM libraries (\eg ORB-SLAM, VINS-Mono, OKVIS, ROVIO) 
by enabling mesh reconstruction and semantic labeling in 3D.
\Kimera is designed with modularity in mind and has four key components: a visual-inertial odometry (VIO)
module for fast and accurate state estimation, a robust pose graph optimizer for global trajectory estimation, a lightweight 3D mesher  module for fast mesh reconstruction,
 and a dense 3D metric-semantic reconstruction module.
The modules can be run in isolation or in combination, hence \Kimera can easily fall back to a state-of-the-art  
VIO or a full SLAM system. 
\Kimera runs in real-time on a CPU and produces a 3D metric-semantic mesh from 
semantically labeled images, which 
 can be obtained by modern deep learning methods. %
We hope that the flexibility, computational efficiency, robustness, and accuracy afforded by \Kimera will 
build a solid basis for future metric-semantic SLAM and perception research, and 
will allow researchers across multiple areas
 (\eg VIO, SLAM, 3D reconstruction, segmentation) to benchmark and prototype their 
 own efforts without having to start from scratch.
\end{abstract}

\vspace{-1mm}
\section*{Supplementary Material} 
\vspace{-1mm}
\centerline{\KimeraUrl}
\centerline{\VideoUrl}

 \newpage

\section{Introduction}\label{sec:introduction}
  Metric-semantic understanding is the capability to simultaneously estimate
the 3D geometry of a scene and attach a semantic label to objects and structures (\eg tables, walls).
  Geometric information is critical for robots to navigate safely and to manipulate objects, while semantic information provides the ideal level of abstraction for a robot to understand and execute human instructions
  (\eg``bring me a cup of coffee'', ``exit from the red door'') and to provide humans with models of the environment that are easy to understand.

  Despite the unprecedented progress in \textit{geometric reconstruction} (\eg SLAM~\cite{Cadena16tro-SLAMsurvey}, Structure from Motion~\cite{Enqvist11iccv}, and Multi-View Stereo~\cite{Schoeps17cvpr}) and deep-learning-based \textit{semantic segmentation} (\eg\cite{GarciaGarcia17arxiv,Krizhevsky12cvpr-deepNets,Redmon17cvpr-yolo9000,Ren15nips-RCNN,He17iccv-maskRCNN,Hu18cvpr-maskXRCNN,Badrinarayanan15pami-segnet}),
  research in these two fields has traditionally proceeded in isolation. However, recently there has been a growing interest towards research and applications at the intersection of these areas~\cite{Bao11cvpr,Cadena16tro-SLAMsurvey,Bowman17icra,Hackel17arxiv-semantic3d,Grinvald19ral-voxbloxpp,Zheng19arxiv-metricSemantic}.

  This growing interest motivated us to create and release \emph{\Kimera}, a library for metric-semantic localization and mapping that combines the state of the art in geometric and semantic understanding into a modern  perception library.
  Contrary to related efforts targeting visual-inertial odometry (VIO) and SLAM,
  we combine visual-inertial SLAM, mesh reconstruction, and semantic understanding.
  Our effort also complements approaches at the
  boundary between metric and semantic understanding in several aspects. %
  First, while existing efforts %
  focus on RGB-D sensing,
  \Kimera uses \emph{visual} (RGB) and \emph{inertial} sensing, which works well in a broader variety of (indoor and outdoor) environments.
  Second, while related works~\cite{Salas-Moreno13cvpr,McCormac17icra-semanticFusion,Whelan15rss-elasticfusion} require a GPU for 3D mapping, we provide a \emph{fast, lightweight, and scalable} CPU-based
  solution.
  Finally, we focus on \emph{robustness}: we include state-of-the-art outlier rejection methods to ensure that \Kimera executes robustly and with minimal parameter tuning across a variety of scenarios,
from real benchmarking datasets~\cite{Burri16ijrr-eurocDataset} to photo-realistic
  simulations~\cite{SayreMcCord18icra-droneSystem,Guerra19arxiv-flightGoggles}.

\begin{table}
  \vspace{0.7mm}
\scriptsize{
  \center
  \setlength\tabcolsep{4pt}
    \begin{tabular}{c||c|c|c|c}
    Method                              
    & \makecell{Sensors} & \makecell{Back-end}  &  \makecell{Geometry}  &  \makecell{Sema-\\ntics} \\ %
    \hline
    ORB-SLAM~\cite{Mur-Artal15tro}      &   mono                       &   \gtwoo                  &     points 
    &   \myNo                          \\%
    \hline
    DSO~\cite{Engel18pami-DSO}          &    mono                   &   \gtwoo                  &     points
    &  \myNo                     \\%
    \hline
    VINS-mono~\cite{Qin18tro-vinsmono}  &   mono/IMU      &   Ceres    &     points 
    &   \myNo     \\%
    \hline
    VINS-Fusion~\cite{Qin19arxiv-VINS-Fusion-odometry}  &   \makecell{mono/Stereo/IMU}  & Ceres   &  points   
    &   \myNo      \\%
    \hline
    ROVIOLI~\cite{Schneider18ral-maplab} &   stereo/IMU                   &  EKF                &     points
    &  \myNo                      \\%
    \hline
    ElasticFusion~\cite{Whelan15rss-elasticfusion}       &   RGB-D                  &  alternation               &    surfels
    &  \myNo                      \\%
    \hline
    Voxblox~\cite{Oleynikova2017iros-voxblox}        &   RGB-D                   &   \cite{Schneider18ral-maplab}                &     TSDF
    &   \myNo                    \\%
    \hline
    \hline
    SLAM++~\cite{Salas-Moreno13cvpr}          &   RGB-D                 &   alternation                  &     objects
    &  \myYes                      \\%
    \hline
    SemanticFusion~\cite{McCormac17icra-semanticFusion}   &   RGB-D                   &  \cite{Whelan15rss-elasticfusion}               &   surfels
    &  \myYes                      \\%
    \hline
    Mask-fusion~\cite{Runz18ismar-maskfusion}   &   RGB-D                   &   \cite{Keller133dv}                  &    surfels
    &  \myYes                     \\%
    \hline
    SegMap~\cite{Renaud18rss-segMap}   &   lidar                  &   GTSAM               &     points/segments  
    &  \myYes                      \\%
    \hline
    XIVO~\cite{Dong17cvpr-XVIO}         &  mono/IMU                   &   EKF                    &    objects 
    &  \myYes                       \\%
    \hline
    Voxblox++~\cite{Grinvald19ral-voxbloxpp}        &   RGB-D                   &   \cite{Schneider18ral-maplab}                &     TSDF
    &   \myYes                    \\%
    \hline
    \hline
    {\bf\Kimera} &    {\bf mono/stereo/IMU}     &   {\bf GTSAM}                   &     {\bf mesh/TSDF}
    &  {\bf\myYes}                   \\%
    \hline
    \end{tabular} 
    \caption{Related \emph{open-source} libraries for visual and visual-inertial SLAM (top) and metric-semantic reconstruction (bottom).  
    \label{tab:related_work}\vspace{-10mm}}
}
\end{table}

\myparagraph{Related Work}
\isExtended{We review work on metric-semantic understanding, while we refer the reader to Table~\ref{tab:related_work} for a visual
comparison against existing VIO and visual-SLAM systems, and to~\cite{Cadena16tro-SLAMsurvey} for a broader review on SLAM.}{
  We refer the reader to Table~\ref{tab:related_work} for a visual
comparison against existing VIO and visual-SLAM systems, and to~\cite{Cadena16tro-SLAMsurvey} for a broader review on SLAM.
}
While early work on metric-semantic understanding~\cite{Bao11cvpr,Brostow08eccv} were designed for offline processing,
recent years have seen a surge of interest towards \emph{real-time} metric-semantic mapping, %
 triggered by pioneering works such as SLAM++~\cite{Salas-Moreno13cvpr}.
 Most of these works (i) rely on RGB-D cameras, (ii) use GPU processing, (iii) %
 alternate tracking and mapping (``alternation'' in Table~\ref{tab:related_work}),
 and (iv) use voxel-based (\eg Truncated Signed Distance Function, TSDF), surfel, or object representations.
 Examples include SemanticFusion~\cite{McCormac17icra-semanticFusion},
  the approach of Zheng~\etal\cite{Zheng19arxiv-metricSemantic}, Tateno~\etal~\cite{Tateno15iros-metricSemantic},
  and Li~\etal~\cite{Li16iros-metricSemantic},
  Fusion++~\cite{McCormac183dv-fusion++},
  Mask-fusion~\cite{Runz18ismar-maskfusion},
  Co-fusion~\cite{Runz17icra-cofusion},
  and MID-Fusion~\cite{Xu19icra-midFusion}.
  Recent work investigates CPU-based approaches, \eg Wald~\etal~\cite{Wald18ral-metricSemantic},
   PanopticFusion~\cite{Narita19arxiv-metricSemantic}, and Voxblox++~\cite{Grinvald19ral-voxbloxpp}; these also rely on RGB-D sensing.
   A sparser set of contributions address other sensing modalities, including monocular cameras
   (\eg CNN-SLAM~\cite{Tateno17cvpr-CNN-SLAM}, VSO~\cite{Lianos18eccv-VSO}, VITAMIN-E~\cite{Yokozuka19arxiv-vitamine}, XIVO~\cite{Dong17cvpr-XVIO}) and lidar (\eg SemanticKitti~\cite{Behley19iccv-semanticKitti}, SegMap~\cite{Renaud18rss-segMap}).
 \isExtended{XIVO~\cite{Dong17cvpr-XVIO} and Voxblox++~\cite{Grinvald19ral-voxbloxpp} are the closest to our proposal.
 XIVO~\cite{Dong17cvpr-XVIO} is an EKF-based visual-inertial approach and produces a set of objects akin to landmark-based SLAM.
Voxblox++~\cite{Grinvald19ral-voxbloxpp} relies on RGB-D sensing and builds on
 Maplab~\cite{Schneider18ral-maplab};
the latter is mainly designed for offline operation, but can be used online using an EKF-based VIO approach
 (ROVIOLI~\cite{Schneider18ral-maplab}). Contrarily to these works, \Kimera (i) provides a highly-accurate
optimization-based VIO approach, (ii) uses a robust and versatile pose graph optimizer,
and (iii) provides multiple lightweight implementations for mesh reconstruction.}{
 XIVO~\cite{Dong17cvpr-XVIO} and Voxblox++~\cite{Grinvald19ral-voxbloxpp} are the closest to our proposal.
 XIVO~\cite{Dong17cvpr-XVIO} is an EKF-based visual-inertial approach and produces an object-based map.
Voxblox++~\cite{Grinvald19ral-voxbloxpp} relies on RGB-D sensing\Toni{, wheel odometry, and pre-built maps} \Toni{\st{and on} using maplab~\cite{Schneider18ral-maplab}}
to obtain visual-inertial pose estimates.
 Contrary to these works, \Kimera (i) provides a highly-accurate real-time
optimization-based VIO, (ii) uses a robust and versatile pose graph optimizer,
and (iii) provides a lightweight mesh reconstruction.
}
\myparagraph{Contribution}
 We release
 \emph{\Kimera}, \myClaim open-source C++ library
 that uses visual-inertial sensing to estimate the state of the robot and build a lightweight
 metric-semantic mesh model of the environment. %
 The name \Kimera stems from the hybrid nature of our library, that unifies state-of-the-art
 efforts across research areas, including VIO, pose graph optimization (PGO), mesh reconstruction, and 3D semantic segmentation.
 \Kimera includes four key modules:

 \begin{itemize}%
  \item {\bf \KimeraVIO}: a VIO module for fast and accurate IMU-rate state estimation.
    At its core, \KimeraVIO features a GTSAM-based VIO approach~\cite{gtsam},
     using IMU-preintegration and structureless vision factors~\cite{Forster17tro}, and
     achieves top performance on the \Euroc dataset~\cite{Burri16ijrr-eurocDataset};

  \item {\bf \KimeraRPGO}: a robust pose graph optimization (RPGO) method that capitalizes on modern techniques for outlier rejection~\cite{Mangelson18icra}. \KimeraRPGO adds a robustness layer that avoids SLAM failures due to   perceptual aliasing, and relieves the user from time-consuming parameter tuning; %

  \item {\bf \KimeraMesher}:
     a module that computes
     a fast per-frame and multi-frame regularized 3D mesh to support obstacle avoidance.
     The mesher builds on previous algorithms by the authors and other groups~\cite{Rosinol19icra-mesh, Greene17flame-iccv, Teixeira16iros, Yokozuka19arxiv-vitamine};

 \item {\bf \KimeraSemantics}:
 a module that builds a slower-but-more-accurate global %
     3D mesh using a volumetric approach~\cite{Oleynikova2017iros-voxblox}, and
     semantically annotates the 3D mesh using 2D pixel-wise semantic segmentation.
 \end{itemize}

\Kimera can work both with offline datasets or online using the Robot Operating System (ROS) \cite{Quigley09icra-ros}.
It runs in real-time on a CPU and provides useful debugging and visualization tools. Moreover, it is modular and
allows replacing each module or executing them in isolation.
For instance, it can fall back to a VIO solution or can simply estimate a geometric mesh if the semantic labels are not available.
\isExtended{The source code of \Kimera is available at \KimeraUrl.}{}

\begin{figure*}[t!]
    \centering
    \includegraphics[trim=0.2cm 0.4cm 1mm 6mm, clip, width=\textwidth]{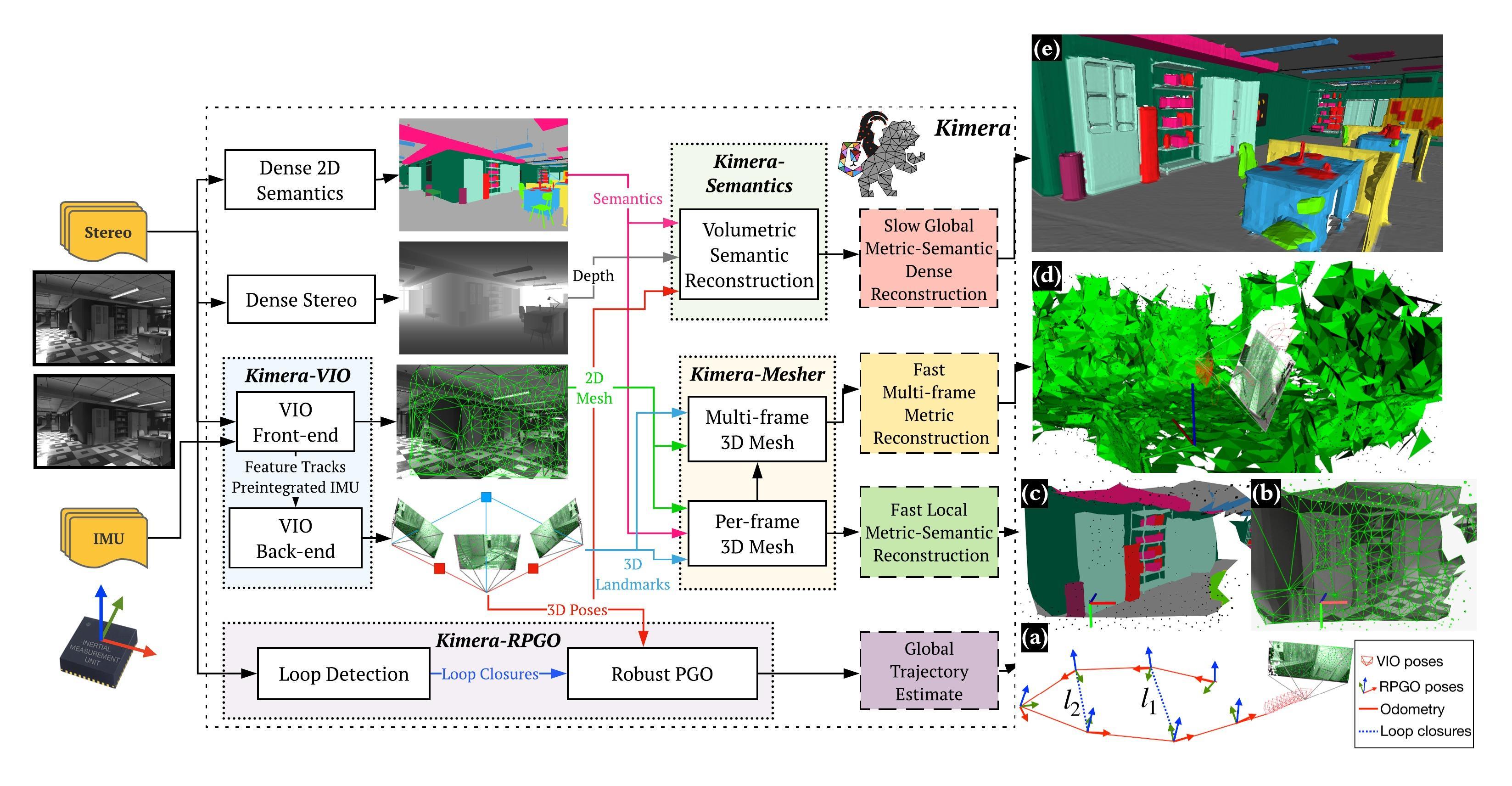}\vspace{-2mm}
    \caption{\Kimera's architecture. \Kimera uses images and IMU data as input (shown on the left) 
    	and outputs (a) pose estimates and (b-e) multiple metric-semantic reconstructions. 
    	\Kimera has 4 key modules: \KimeraVIO, \KimeraRPGO, \KimeraMesher, \KimeraSemantics.
    \label{fig:kimera_diagram}  \vspace{-5mm}}    
\end{figure*} 
\section{\Kimera}
\label{sec:methodology}

\Cref{fig:kimera_diagram} shows \Kimera's architecture. 
\Kimera takes stereo frames and high-rate inertial measurements as input and returns 
(i) a highly accurate state estimate at IMU rate, 
(ii) a globally-consistent trajectory estimate, and
(iii) multiple meshes of the environment, including a fast local mesh and a global 
semantically annotated mesh. 
\Kimera is heavily parallelized and uses four threads to accommodate inputs and outputs at different rates 
(\eg IMU, frames, keyframes). Here we describe the architecture \emph{by threads}, while the description of each module is given in the following sections.

The first thread includes the \KimeraVIO front-end (Section~\ref{sec:vio}) that takes stereo images and IMU data and outputs feature tracks and preintegrated IMU measurements. The front-end also publishes IMU-rate state estimates.
 The second thread includes (i) the \KimeraVIO back-end, that outputs optimized state estimates, and 
 (ii) \KimeraMesher (Section~\ref{sec:mesher}), that 
  computes low-latency ($<\!\perFrameMeshLatency$) per-frame and multi-frame 3D meshes. 
  These two threads allow creating the per-frame mesh in~\Cref{fig:kimera_diagram}(b) (which can also come with semantic labels as in~\Cref{fig:kimera_diagram}(c)), 
  as well as the multi-frame mesh in~\Cref{fig:kimera_diagram}(d).
 The last two threads operate at slower rate and are designed to support low-frequency functionalities, such as path planning.
  The third thread includes \KimeraRPGO (Section~\ref{sec:rpgo}), a robust PGO implementation that detects loop closures, 
  rejects outliers, and estimates a globally consistent trajectory (\Cref{fig:kimera_diagram}(a)). 
  The last thread includes \KimeraSemantics (Section~\ref{sec:semantic}), that uses dense stereo and 2D semantic labels to obtain a refined metric-semantic mesh, using \Toni{\st{the pose estimates by} \KimeraVIO's pose estimates}.
\subsection{\KimeraVIO: Visual-Inertial Odometry Module}
\label{sec:vio}

\KimeraVIO implements the keyframe-based maximum-a-posteriori visual-inertial estimator presented in~\cite{Forster17tro}.
 In our implementation, the estimator can perform both \emph{full} smoothing or \emph{fixed-lag} smoothing, depending on the 
 specified time horizon; we typically use the latter \Toni{\st{since it ensures bounded} to bound the} estimation time. 
 We also extend~\cite{Forster17tro} to work with both monocular and stereo frames. %
 \KimeraVIO includes a (visual and inertial) front-end which is in charge of processing the raw sensor data, and a 
 back-end, that fuses the processed measurements to obtain an estimate of the state of the sensors (\ie pose, velocity, and sensor biases). %

\subsubsection{VIO Front-end} 
Our IMU front-end performs on-manifold preintegration~\cite{Forster17tro} to obtain compact preintegrated measurements of the 
relative state between two consecutive keyframes from raw IMU data. 
The vision front-end detects Shi-Tomasi corners~\cite{Shi94}, tracks them across frames using the Lukas-Kanade tracker~\cite{Bouguet00-lktracking}, finds left-right stereo matches, and performs geometric verification \Toni{\st{Would be nice to say that these can be changed easily...}}. 
We perform both mono(cular) verification using 5-point RANSAC~\cite{Nister04pami} and stereo verification using 
 3-point RANSAC~\cite{Horn87josa}; the code also offers the option to use the IMU rotation and perform mono and stereo verification using  2-point~\cite{Kneip11bmvc} and  1-point RANSAC, respectively.  
Feature detection, stereo matching, and geometric verification are executed at each \emph{keyframe}, while 
we only track features at intermediate \emph{frames}.
  
\subsubsection{VIO Back-end} 
At each keyframe, preintegrated IMU and visual measurements are added to a fixed-lag smoother (a factor graph) which constitutes our VIO back-end. 
We use the preintegrated IMU model and  the structureless vision model of~\cite{Forster17tro}.
The factor graph is solved using iSAM2~\cite{Kaess12ijrr} in GTSAM~\cite{Dellaert12tr}.
At each iSAM2 iteration, 
the structureless vision model estimates the 3D position of the observed features using DLT~\cite{Hartley04book}
 and analytically eliminates the corresponding 3D points from the VIO state~\cite{Carlone14icra-smartFactors}. 
Before elimination, degenerate points (\ie points behind the camera or without enough parallax for triangulation) and outliers (\ie points with large reprojection error) are removed, providing an extra robustness layer. 
Finally, states that fall out of the 
smoothing horizon are marginalized out using GTSAM. %

\isExtended{
	\newcommand{\figRPGO}{Fig.~\ref{fig:RPGO}\xspace}
}{
	\newcommand{\figRPGO}{\Cref{fig:kimera_diagram}(a)\xspace}
}

\subsection{\KimeraRPGO: Robust Pose Graph Optimization Module}
\label{sec:rpgo}

\KimeraRPGO is in charge of (i) detecting loop closures between the current and past keyframes, 
and
(ii) computing globally consistent keyframe poses using robust PGO. 

\subsubsection{Loop Closure Detection}
The loop closure detection relies on the DBoW2 library~\cite{Galvez12tro-dbow} and uses a bag-of-word representation to quickly detect putative loop closures. 
For each putative loop closure, we reject outlier loop closures using mono and stereo geometric verification (as described in Section~\ref{sec:vio}), and pass the remaining loop closures to the robust PGO solver. 
Note that the resulting loop closures can still contain outliers due to perceptual aliasing (\eg two identical rooms on 
different floors of a building).

\subsubsection{Robust PGO}
This module is implemented in GTSAM, and includes a modern outlier rejection method, \emph{Incremental 
Consistent Measurement Set Maximization} (PCM)~\cite{Mangelson18icra}, that we 
tailor to a single-robot and online setup. We store separately the odometry edges (produced by \KimeraVIO) and the loop closures (produced by the loop closure detection); each time the PGO is executed, we first 
select the largest set of consistent loop closures using a modified version of PCM, and then execute GTSAM on the pose graph including the odometry and the consistent loop closures.

\isExtended{
\begin{figure}[h!]
  \centering
  \includegraphics[width=\columnwidth]{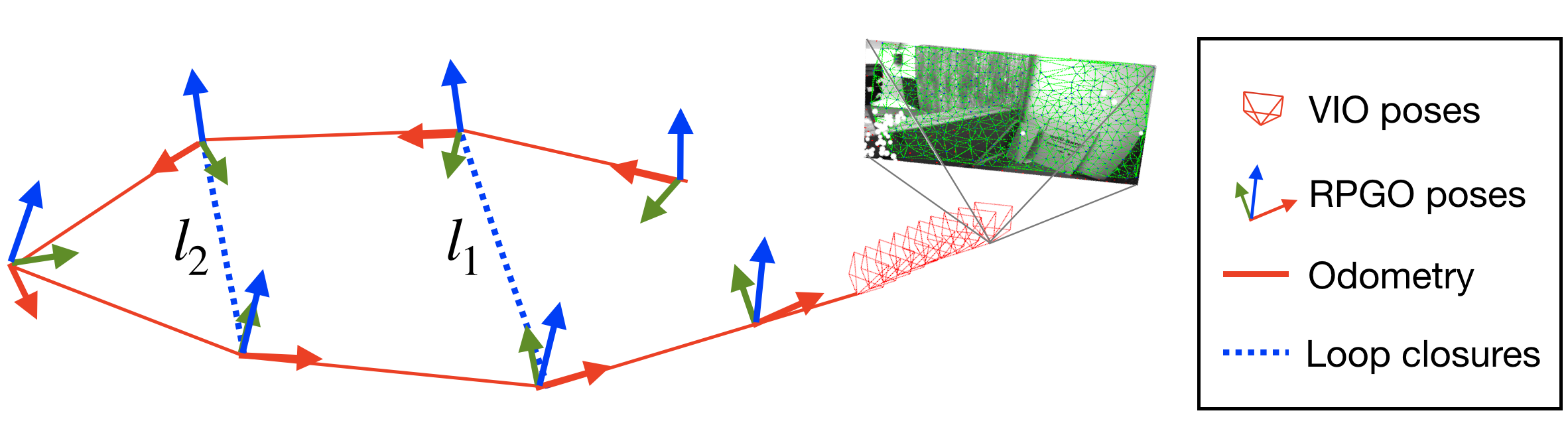}
  \caption{\KimeraRPGO detects visual loop closures, rejects spurious loop closures, and estimates a globally consistent trajectory. \vspace{-5mm} \label{fig:RPGO}}
\end{figure}
}{}

 PCM is designed for the multi-robot case and only checks that inter-robot loop closures are consistent. 
 We developed a C++ implementation of PCM that (i) adds an \emph{odometry consistency check} on the loop closures and 
 (ii) \emph{incrementally} updates the set of consistent measurements to enable online operation.
 The odometry check verifies that each loop closure (\eg $l_1$ in \figRPGO) is consistent with the odometry (in red in the figure): 
 in the absence of noise, the poses along the cycle formed by the odometry and the loop $l_1$ must compose to the identity. As in PCM, we flag as outliers loops for which the error accumulated along the cycle is not consistent with the measurement noise using a Chi-squared test. If a loop detected at the current time $t$ passes the odometry check, we test if it is pairwise consistent with previous loop closures as in~\cite{Mangelson18icra} (\eg check if loops $l_1$ and $l_2$ in \figRPGO are consistent with each other). 
 \Toni{While} PCM~\cite{Mangelson18icra} builds an adjacency matrix $\MA \in \Real^{L \times L}$ \Toni{from scratch\st{, where $L$ is the number of detected loop closures,}} to keep track of pairwise-consistent loops \Toni{(where $L$ is the number of detected loop closures)\st{. To}, we} enable online operation \Toni{\st{, rather than} by} building the matrix $\MA$ \Toni{\st{from scratch,} incrementally.} 
 \Toni{\st{each} Each} time a new loop is detected, we\Toni{\st{only}} add a row and column \Toni{\st{to the matrix} to the matrix $\MA$} and only test the new loop against the previous ones.
  Finally, we use the fast maximum clique implementation of~\cite{Pattabiraman15im-maxClique} to compute the largest set of consistent loop closures.
   The set of consistent measurements are added to the pose graph (together with the odometry) and optimized using \Toni{\st{the Gauss-Newton method in GTSAM} Gauss-Newton}.
\subsection{\KimeraMesher: 3D Mesh Reconstruction}
\label{sec:mesher}

\KimeraMesher can quickly generate two types of 3D meshes: (i) a per-frame 3D mesh, and (ii) a multi-frame 3D mesh spanning
 the keyframes in the VIO fixed-lag smoother. 

\subsubsection{Per-frame mesh} 
 As in \cite{Rosinol19icra-mesh}, we first perform a 2D Delaunay triangulation over the successfully tracked 2D features (generated by the VIO front-end) in the current keyframe. %
Then, we back-project the 2D Delaunay triangulation to generate a 3D mesh (\cref{fig:kimera_diagram}(b)),
 using the 3D point estimates from the VIO back-end. 
 While the per-frame mesh is designed to provide low-latency obstacle detection, we 
 also provide the option to semantically label the resulting mesh, by texturing the mesh with 2D labels (\cref{fig:kimera_diagram}(c)).

\subsubsection{Multi-frame mesh} 
 The multi-frame mesh fuses the per-frame meshes collected over the VIO receding horizon into a single mesh \Toni{\st{and regularizes planar surfaces}} (\cref{fig:kimera_diagram}(d)). 
Both~\Toni{\st{the}} per-frame and multi-frame 3D meshes are encoded as a list of vertex positions, together with a list of triplets of vertex IDs to describe the triangular faces.
Assuming we already have a multi-frame mesh at time $t-1$, for each new per-frame 3D mesh that we generate (at time $t$), we loop over its vertices and triplets and add vertices and triplets that are in the per-frame mesh but are missing in the multi-frame one.
Then we loop over the multi-frame mesh vertices and update their 3D position according to the latest VIO back-end estimates.
Finally, we remove vertices and triplets corresponding to old features observed outside the VIO time horizon.
The result is an up-to-date 3D mesh spanning the keyframes in the current VIO time horizon. 
If planar surfaces are detected in the mesh, \emph{regularity factors}~\cite{Rosinol19icra-mesh} are added to the VIO back-end, which results in a tight coupling between VIO and mesh regularization, see~\cite{Rosinol19icra-mesh} for further details.  
\subsection{\KimeraSemantics: Metric-Semantic Segmentation}
\label{sec:semantic}

We adapt the \textit{bundled raycasting} technique introduced in \cite{Oleynikova2017iros-voxblox} to 
(i) build an accurate global 3D mesh (covering the entire trajectory), 
and (ii) semantically annotate the mesh. 

\subsubsection{Global mesh}
\mbox{Our implementation builds on Voxblox} \cite{Oleynikova2017iros-voxblox} and 
uses a voxel-based (TSDF) model to filter out noise and extract the global mesh.
At each keyframe,  we use dense stereo (semi-global matching~\cite{Hirschmuller08pami}) to obtain a 3D point cloud from the current stereo pair. Then we apply bundled raycasting using Voxblox~\cite{Oleynikova2017iros-voxblox}, using the ``fast'' option discussed in~\cite{Oleynikova2017iros-voxblox}. This process is repeated at each keyframe and produces a TSFD, from which a mesh is extracted using marching cubes~\cite{Lorensen87siggraph-marchingCubes}. 

\subsubsection{Semantic annotation}
\KimeraSemantics uses 2D semantically labeled images (produced at each keyframe) to semantically annotate the global mesh;
the 2D semantic labels can be obtained using 
off-the-shelf tools for pixel-level 2D semantic segmentation, \eg deep neural networks~\cite{Huang19arxiv,Zhang19bmvc,Chen17pami,Zhao17cvpr,Yang18eccv,Paszke16arxiv,Ren15nips-RCNN,He17iccv-maskRCNN,Hu18cvpr-maskXRCNN} or classical MRF-based approaches~\cite{Hu19ral-fuses}.
To this end, during the bundled raycasting, we also propagate the semantic labels. 
 Using the 2D semantic segmentation, we attach a label to each 3D point produced by the dense stereo.
 Then, for each bundle of rays in the {bundled raycasting}, %
 we build a vector of label probabilities from the frequency of the observed labels in the bundle.
We then propagate this information along the ray only within the TSDF truncation distance (\ie near the surface) to spare computation. %
In other words, we spare the computational effort of updating probabilities for the ``empty'' label.
While traversing the voxels along the ray, we use a Bayesian update to update the label probabilities at each voxel, similar to~\cite{McCormac17icra-semanticFusion}.
After \Toni{\st{this}} bundled semantic raycasting, each voxel has a vector of label probabilities, from which we extract the 
most likely label. The metric-semantic mesh is finally extracted using marching cubes~\cite{Lorensen87siggraph-marchingCubes}.
The resulting mesh is significantly more accurate than the multi-frame mesh of Section~\ref{sec:mesher}, but it is slower to compute ($\approx \globalMeshLatency$, see Section~\ref{ssec:timing_performance}). 	%
\subsection{Debugging Tools}
\label{sec:debugging}
 
While we limit the discussion for space reasons, 
 it is worth mentioning that \Kimera also provides an open-source suite of evaluation tools  for debugging, visualization, and benchmarking of VIO, SLAM, and
  metric-semantic reconstruction.
 \Kimera includes a Continuous Integration server (Jenkins) that asserts the quality of the code (compilation, unit tests), but also automatically evaluates \KimeraVIO and \KimeraRPGO on the \Euroc's datasets using \emph{evo}~\cite{Grupp17evo}. Moreover, we provide Jupyter Notebooks to visualize intermediate VIO statistics (\eg quality of the feature tracks,   IMU preintegration errors), as well as to automatically assess the quality of the 3D reconstruction using Open3D~\cite{Zhou18arxiv-open3D}.

\section{Experimental Evaluation}
\label{sec:results}

\cref{ssec:localization_performance} shows that (i) \Kimera attains state-of-the-art state estimation performance 
and (ii) our robust PGO relieves the user from time-consuming parameter tuning. 
\cref{ssec:geometric_performance} demonstrates \Kimera's 3D mesh reconstruction on \Euroc, using the subset of scenes providing a ground-truth point cloud.
\cref{ssec:semantic_performance} inspects \Kimera's 3D metric-semantic reconstruction using 
a  photo-realistic simulator (see video attachment), which provides ground-truth 3D semantics. %
Finally, \cref{ssec:timing_performance} highlights 
\Kimera's real-time performance and analyzes the runtime of each module.

\isExtended{

\begin{figure}[t]
  \centering
  \includegraphics[trim={0 0cm 0 -0.5cm}, clip, width=0.49\columnwidth]{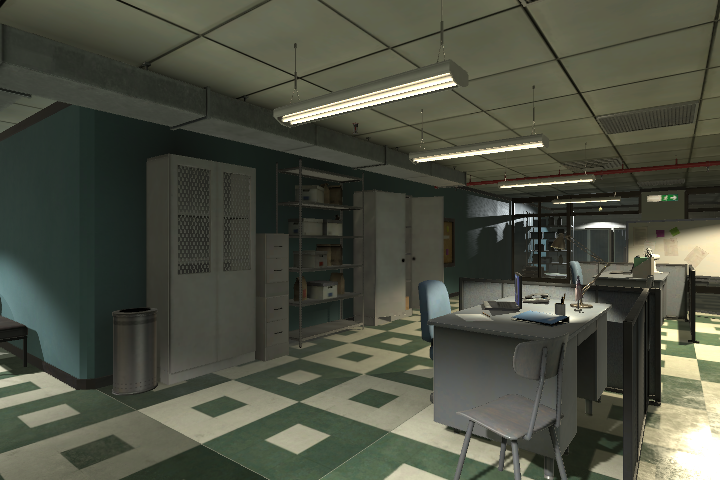}
  \includegraphics[trim={0 0cm 0 -0.5cm}, clip, width=0.49\columnwidth]{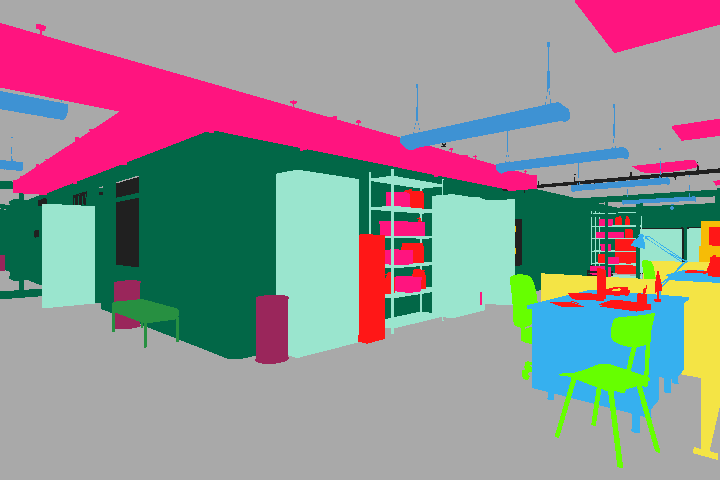}
  \caption{Simulator: RGB image and 2D semantic segmentation.\label{fig:simulator} }
\end{figure} }{} 	%

\subsection{Pose Estimation Performance}
\label{ssec:localization_performance}

\cref{tab:ape_accuracy_comparison_sopa} compares the Root Mean Squared Error (RMSE) of the Absolute Translation Error (ATE) of \KimeraVIO against
state-of-the-art open-source VIO pipelines: OKVIS~\cite{Leutenegger13rss}, MSCKF~\cite{Mourikis07icra},
 ROVIO~\cite{Blosch15iros}, VINS-Mono~\cite{Qin18tro-vinsmono}, \Toni{and SVO-GTSAM~\cite{Forster15rss-imuPreintegration} 
 using the independently reported values in~\cite{Delmerico18icra} and the self-reported values in~\cite{Qin18tro-vinsmono}.
Note that these algorithms use a monocular camera, while we use a stereo camera. 
We align the estimated and ground-truth trajectories using an \SEthree transformation before evaluating the errors. 
Using a $\mathrm{Sim}(3)$ alignment, as in~\cite{Delmerico18icra}, would result in an even smaller error for 
\Kimera: we preferred the \SEthree alignment, since it is more appropriate 
for VIO, where the scale is observable thanks to the IMU.}
\Toni{We group the techniques depending on whether they use fixed-lag smoothing, full smoothing, and loop closures.}
\KimeraVIO and \KimeraRPGO achieve top performance \Toni{across the spectrum.}

\Toni{Furthermore, }\KimeraRPGO ensures robust performance, and is less sensitive to loop closure parameter tuning.
\Toni{\st{To showcase this desirable feature, }}Table~\ref{tab:ape_alpha} shows the PGO accuracy with and without 
 outlier rejection (PCM) for different values of the loop closure threshold $\alpha$ used in DBoW2.
 Small values of $\alpha$ lead to \Toni{\st{many} more} loop closure \Toni{detections}, \Toni{but these are less conservative (more outliers)\st{while for large values ($\alpha=10$) no loop closure is selected 
 and PGO returns the odometric estimate}}.
  Table~\ref{tab:ape_alpha} shows that\Toni{\st{without PCM the choice of $\alpha$ largely influences the PGO accuracy. On the other hand, thanks to PCM, Kimera-RPGO is fairly insensitive to $\alpha$.}, by using PCM, \KimeraRPGO is fairly insensitive to the choice of $\alpha$.}
 The results in \cref{tab:ape_accuracy_comparison_sopa} use $\alpha=0.001$.
\begin{table}[t!]
  \vspace{1mm}
  \centering
  \caption{RMSE of state-of-the-art VIO pipelines (reported from \cite{Delmerico18icra} and~\cite{Qin18tro-vinsmono}) compared to \Kimera, on the \Euroc dataset.
  In \textbf{bold} the best result \Toni{for each category: fixed-lag smoothing, full smoothing, and PGO with loop closure. $\times$ indicates failure.}}
  \vspace{-0.5mm}
  \label{tab:ape_accuracy_comparison_sopa}
  \setlength{\tabcolsep}{3pt}
  \begin{tabularx}{\columnwidth}{*6{Y} || *2{Y} || *2{Y}}
    \toprule
    & \multicolumn{9}{c}{RMSE ATE [m]} \\
    \cmidrule(l){2-10}
    & \multicolumn{5}{c}{Fixed-lag Smoothing} & \multicolumn{2}{c}{Full Smoothing} & \multicolumn{2}{c}{Loop Closure} \\
    \cmidrule(l){2-6} \cmidrule(l){7-8} \cmidrule(l){9-10}
    Seq. & \rotatebox[origin=c]{90}{OKVIS} & \rotatebox[origin=c]{90}{MSCKF} & \rotatebox[origin=c]{90}{ROVIO} & \rotatebox[origin=c]{90}{\specialcell[b]{VINS-\\Mono}} & \rotatebox[origin=c]{90}{\specialcell[b]{\textbf{\!\Kimera-} \\ \textbf{VIO}}}
         & \rotatebox[origin=c]{90}{\specialcell[b]{SVO-\\GTSAM}}   & \rotatebox[origin=c]{90}{\specialcell[b]{\textbf{Kimera-} \\ \textbf{VIO}}}
         & \rotatebox[origin=c]{90}{\specialcell[b]{VINS-\\LC}} & \rotatebox[origin=c]{90}{\specialcell[b]{\textbf{\Kimera-} \\ \textbf{RPGO}}} \\
    \midrule
    MH\_1 & 0.16 & 0.42 & 0.21           & \vinVersion{0.27}{0.15}                    & \textbf{0.11}        & 0.05             & \textbf{0.04}  & \vinVersion{0.07}{0.12}          & \textbf{0.08} \\
    MH\_2 & 0.22 & 0.45 & 0.25           & \vinVersion{0.12}{0.15}                    & \textbf{0.10}        & \textbf{0.03}    & 0.07           & \vinVersion{0.05}{0.12}          & \textbf{0.09} \\
    MH\_3 & 0.24 & 0.23 & 0.25           & \vinVersion{\textbf{0.13}}{0.22}           & \textbf{0.16}        & \textbf{0.12}    & \textbf{0.12}  & \vinVersion{0.08}{0.13}          & \textbf{0.11} \\
    MH\_4 & 0.34 & 0.37 & 0.49           & \vinVersion{0.23}{0.32}                    & \textbf{0.24}        & \textbf{0.13}    & 0.27           & \vinVersion{0.12}{0.18}          & \textbf{0.15} \\
    MH\_5 & 0.47 & 0.48 & 0.52           & \vinVersion{0.35}{\textbf{0.30}}           & 0.35                 & \textbf{0.16}    & 0.20           & \vinVersion{0.09}{\textbf{0.21}} & 0.24          \\
    V1\_1 & 0.09 & 0.34 & 0.10           & \vinVersion{0.07}{0.08}                    & \textbf{0.05}        & 0.07             & \textbf{0.06}  & \vinVersion{0.07}{0.06}          & \textbf{0.05} \\
    V1\_2 & 0.20 & 0.20 & 0.10           & \vinVersion{0.10}{0.11}                    & \textbf{0.08}        & 0.11             & \textbf{0.07}  & \vinVersion{0.06}{\textbf{0.08}} & 0.11          \\
    V1\_3 & 0.24 & 0.67 & 0.14           & \vinVersion{0.13}{0.18}                    & \textbf{0.07}        & $\times$         & \textbf{0.09}  & \vinVersion{0.11}{0.19}          & \textbf{0.12} \\
    V2\_1 & 0.13 & 0.10 & 0.12           & \vinVersion{\textbf{0.08}}{\textbf{0.08}}  & \textbf{0.08}        & \textbf{0.07}    & \textbf{0.07}  & \vinVersion{0.06}{0.08}          & \textbf{0.07} \\
    V2\_2 & 0.16 & 0.16 & 0.14           & \vinVersion{\textbf{0.08}}{0.16}           & \textbf{0.10}        & $\times$         & \textbf{0.09}  & \vinVersion{0.06}{0.16}          & \textbf{0.10} \\
    V2\_3 & 0.29 & 1.13 & \textbf{0.14}  & \vinVersion{0.21}{0.27}                    & {0.21}               & $\times$         & \textbf{0.19}  & \vinVersion{x}{1.39}             & \textbf{0.19} \\
    \bottomrule
  \end{tabularx}
  \vspace{-5mm}
\end{table}

\begin{table}[h!]
  \vspace{1mm}
  \centering
  \caption{RMSE ATE [m] vs. loop closure threshold $\alpha$ (V1\_01).}
  \vspace{-0.5mm}
  \label{tab:ape_alpha}
  \begin{tabular}{cccccc} %
    \toprule
          &  $\alpha\!=\!10$ &  $\alpha\!=\!1$ &  $\alpha\!=\!0.1$ &  $\alpha\!=\!0.01$ & $\alpha\!=\!0.001$\\
    \midrule
    PGO w/o PCM         &0.05 & 0.45 & 1.74 & 1.59 &   1.59 \\
      \KimeraRPGO     & 0.05  & 0.05  & 0.05& 0.045 &  0.049 \\
    \bottomrule
  \end{tabular}
  \vspace{-5.5mm}
\end{table}

\isExtended{
\cref{fig:ate_colormapped} plots \KimeraVIO's estimated trajectory color-coded by its ATE, overlaid on the 
 ground-truth trajectory (dashed line). We generate the plots using~\cite{Grupp17evo}. 

\begin{figure}[htbp]
  \centering
  \includegraphics[trim={0.3cm 0.2cm 1.1cm 0.6cm},clip,width=0.49\columnwidth]{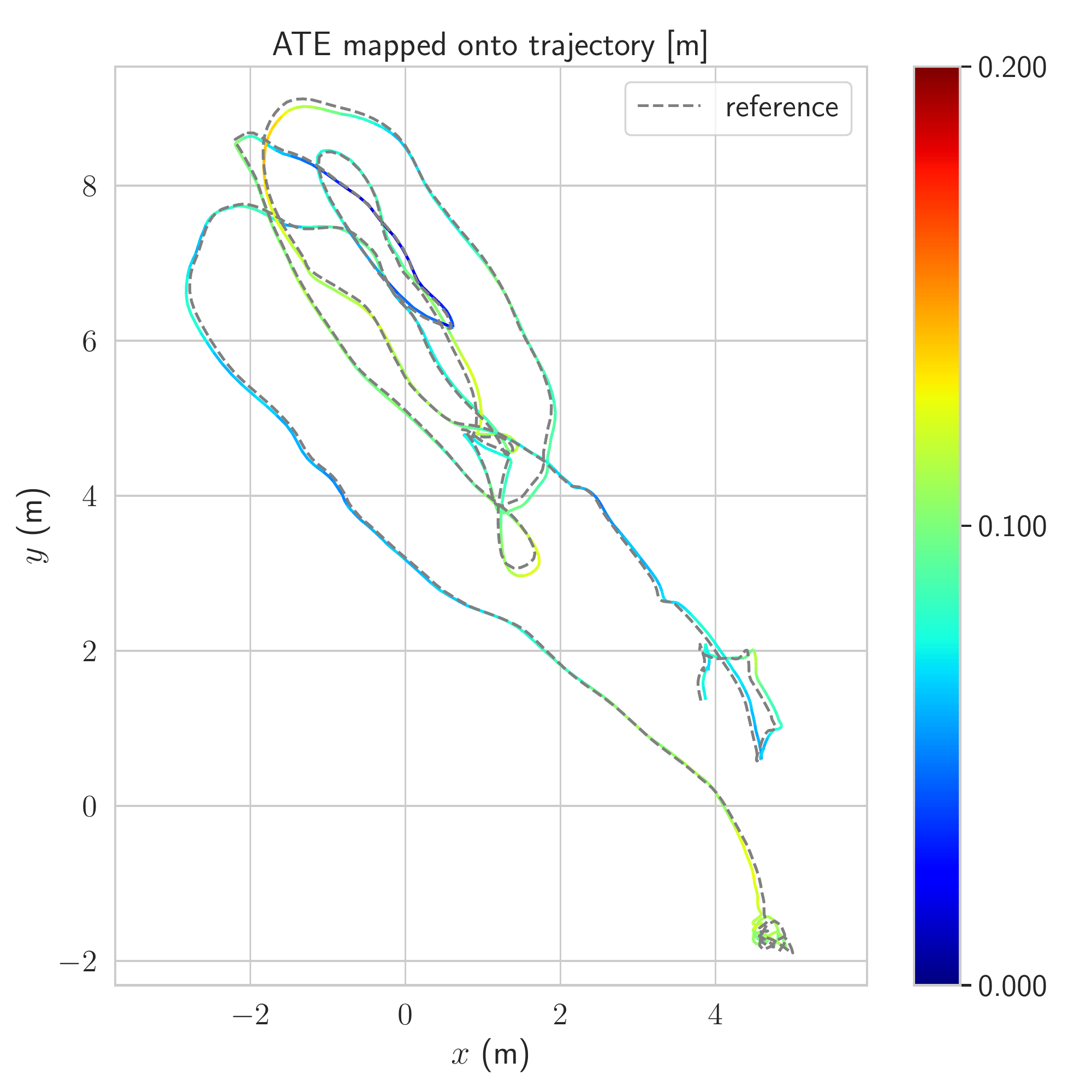}
  \includegraphics[trim={0.3cm 0.2cm 1.1cm 0.6cm},clip,width=0.49\columnwidth]{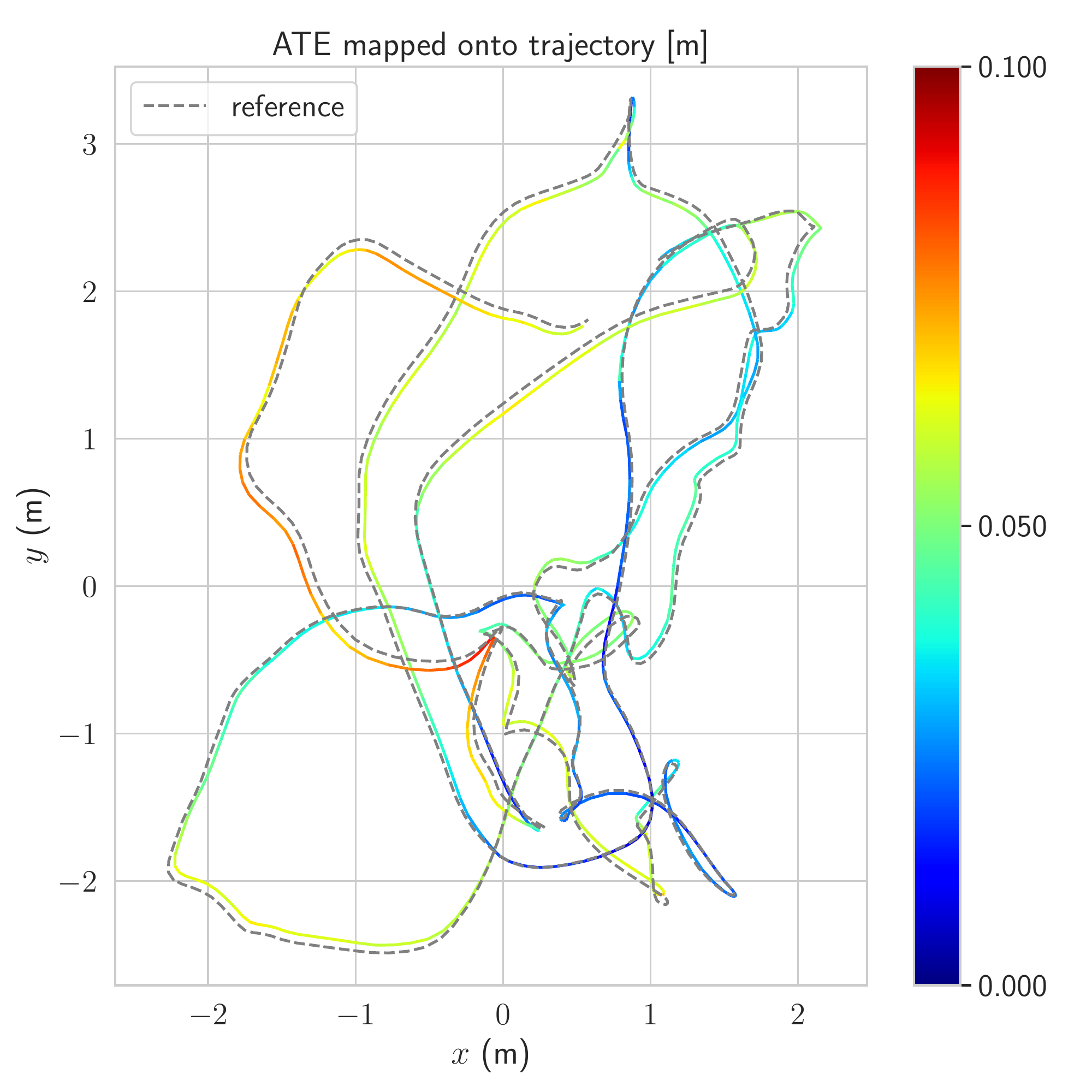}
  \caption{\Euroc  V1\_01 \& MH\_01 dataset. Top-view of \Kimera's estimated trajectory (color-code by their Absolute Translation Error, in meters), and
   ground-truth trajectory (dashed line).
  \label{fig:ate_colormapped}}
\end{figure} }{} 	%

\subsection{Geometric Reconstruction}
\label{ssec:geometric_performance}

\begin{figure}[h!]
\vspace{-3.5mm}
  \centering
  \includegraphics[width=0.7\columnwidth]{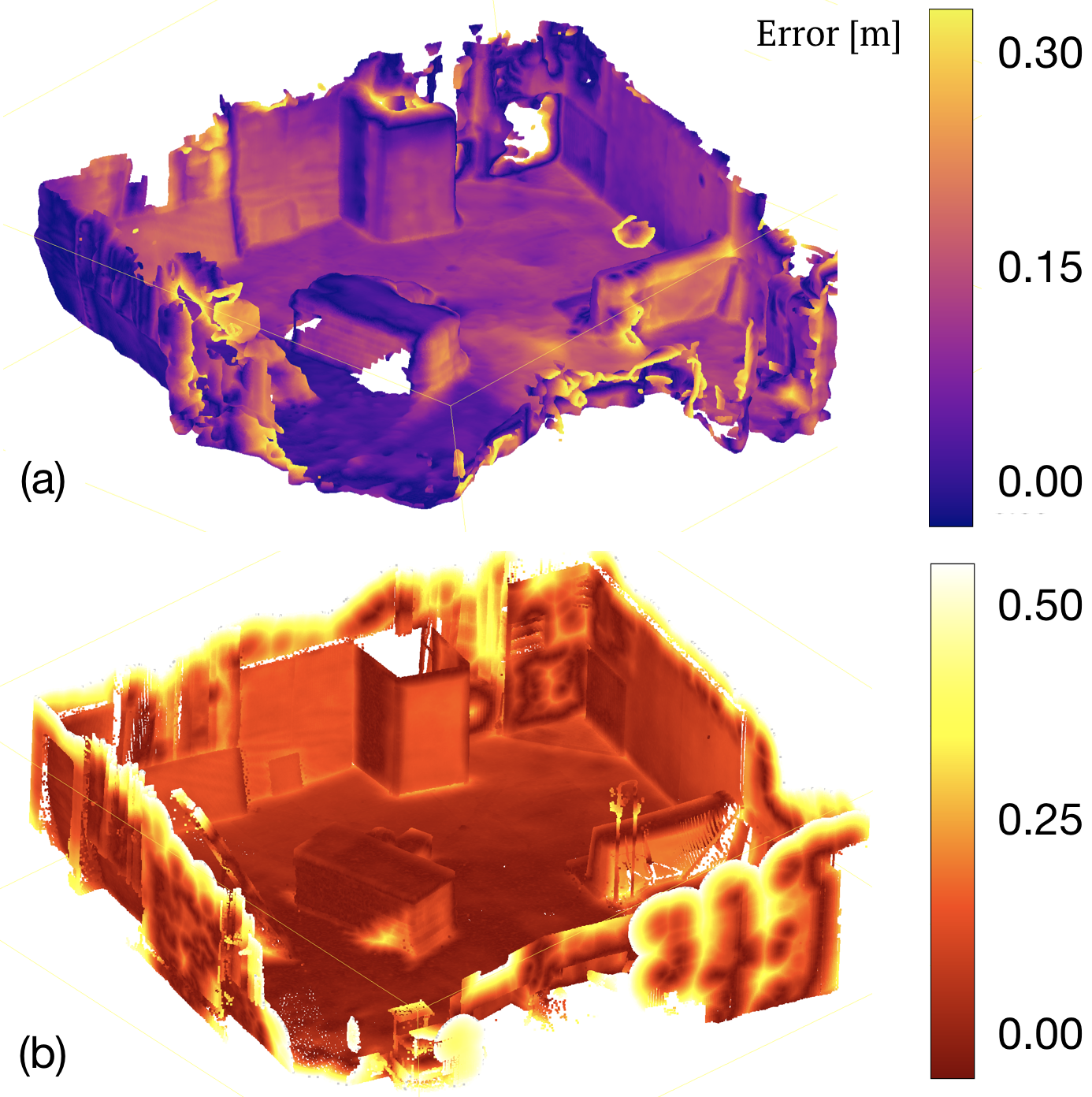}
  \caption{
    (a) \Kimera's 3D mesh color-coded by the distance to the ground-truth point cloud.
    (b) Ground-truth point cloud color-coded by the distance to the estimated  cloud. 
    \Euroc  V1\_01 dataset.\vspace{-3mm}\label{fig:accuracy_mesh}}
\end{figure}

We use the ground truth point cloud available in the \Euroc \texttt{V1} and \texttt{V2} datasets to assess the quality of the 3D meshes
produced by \Kimera.
We evaluate \Toni{\st{our meshes} each mesh} against the ground truth using the \textit{accuracy} and \textit{completeness} metrics as in~\cite[Sec. 4.3]{Rosinol18thesis}:
(i) we compute a point cloud by sampling our mesh with a uniform density of $10^3~\text{points}/\text{m}^2$, 
(ii) we register the estimated and the ground truth clouds with ICP~\cite{Besl92pami} using \emph{CloudCompare}~\cite{CloudCompare}, 
and (iii) we evaluate the average distance from ground truth point cloud to its nearest neighbor in the estimated point cloud (accuracy), and vice-versa (completeness).
\Cref{fig:accuracy_mesh}(a) shows the estimated cloud (corresponding to the global mesh of \KimeraSemantics on V1\_01) color-coded  by the distance to the closest point
in the ground-truth cloud (accuracy); \cref{fig:accuracy_mesh}(b) shows the ground-truth cloud, color-coded with the distance to the closest-point in the estimated cloud (completeness).

Table~\ref{tab:geometric_accuracy} provides a quantitative comparison between the fast multi-frame mesh produced by \KimeraMesher and the slow mesh produced via TSDF by \KimeraSemantics. To obtain a complete mesh from \KimeraMesher we set a large VIO horizon 
(\ie we perform full smoothing).
As expected from \Cref{fig:accuracy_mesh}(a), the global mesh from \KimeraSemantics is very accurate, with an average error of $0.35-0.48$m across datasets. \KimeraMesher produces a \Toni{\st{noisier} more noisy} mesh (up to $24\%$ error increase), 
but %
requires two orders of magnitude less time to compute (see Section~\ref{ssec:timing_performance}).
\begin{table}[htbp]
  \centering
  \vspace{-1mm}
  \caption{Evaluation of \Kimera multi-frame and global meshes' completeness \cite[Sec. 4.3.3]{Rosinol18thesis} with an ICP threshold of $1.0$m.}
  \label{tab:geometric_accuracy}
  \begin{tabularx}{\columnwidth}{l *6{Y}}
    \toprule
    & \multicolumn{2}{c}{\specialcell[b]{RMSE [m]}} &  \multirow{2}{*}{\specialcell[b]{Relative \\ Improvement [\%]}} \\
    \cmidrule(l{2pt}r{2pt}){2-3}
    \multicolumn{1}{c}{Seq.} & \multicolumn{1}{c}{\specialcell[b]{Multi-Frame}} & \multicolumn{1}{c}{\specialcell[b]{Global}} &   \\
    \cmidrule(l{2pt}r{2pt}){2-2} \cmidrule(l{2pt}r{2pt}){3-3} \cmidrule(l{2pt}r{2pt}){4-4}
    V1\_01 & 0.482 & 0.364 & 24.00 \\ 
    V1\_02 & 0.374 & 0.384 & -2.00 \\
    V1\_03 & 0.451 & 0.353 & 21.00 \\
    V2\_01 & 0.465 & 0.480 & -3.00 \\
    V2\_02 & 0.491 & 0.432 & 12.00 \\
    V2\_03 & 0.530 & 0.411 & 22.00 \\
    \bottomrule
  \end{tabularx}
  \vspace{-2mm}
\end{table}

\subsection{Semantic Reconstruction}
\label{ssec:semantic_performance}

To evaluate the accuracy of the metric-semantic reconstruction from \KimeraSemantics, 
we use a photo-realistic 
Unity-based simulator provided by MIT Lincoln Lab, that provides sensor streams (in ROS) and ground truth for both the 
geometry and the semantics of the scene, and has an interface similar to~\cite{SayreMcCord18icra-droneSystem,Guerra19arxiv-flightGoggles}. To avoid biasing the results towards a particular 2D semantic segmentation method, we use ground truth 2D semantic segmentations and we refer the reader to~\cite{Hu19ral-fuses} for potential alternatives.

\begin{figure}[htbp]
  \centering    
  \vspace{-0.5mm} 
  \includegraphics[trim={0cm 2mm 0cm 0cm}, clip, width=0.9\columnwidth]{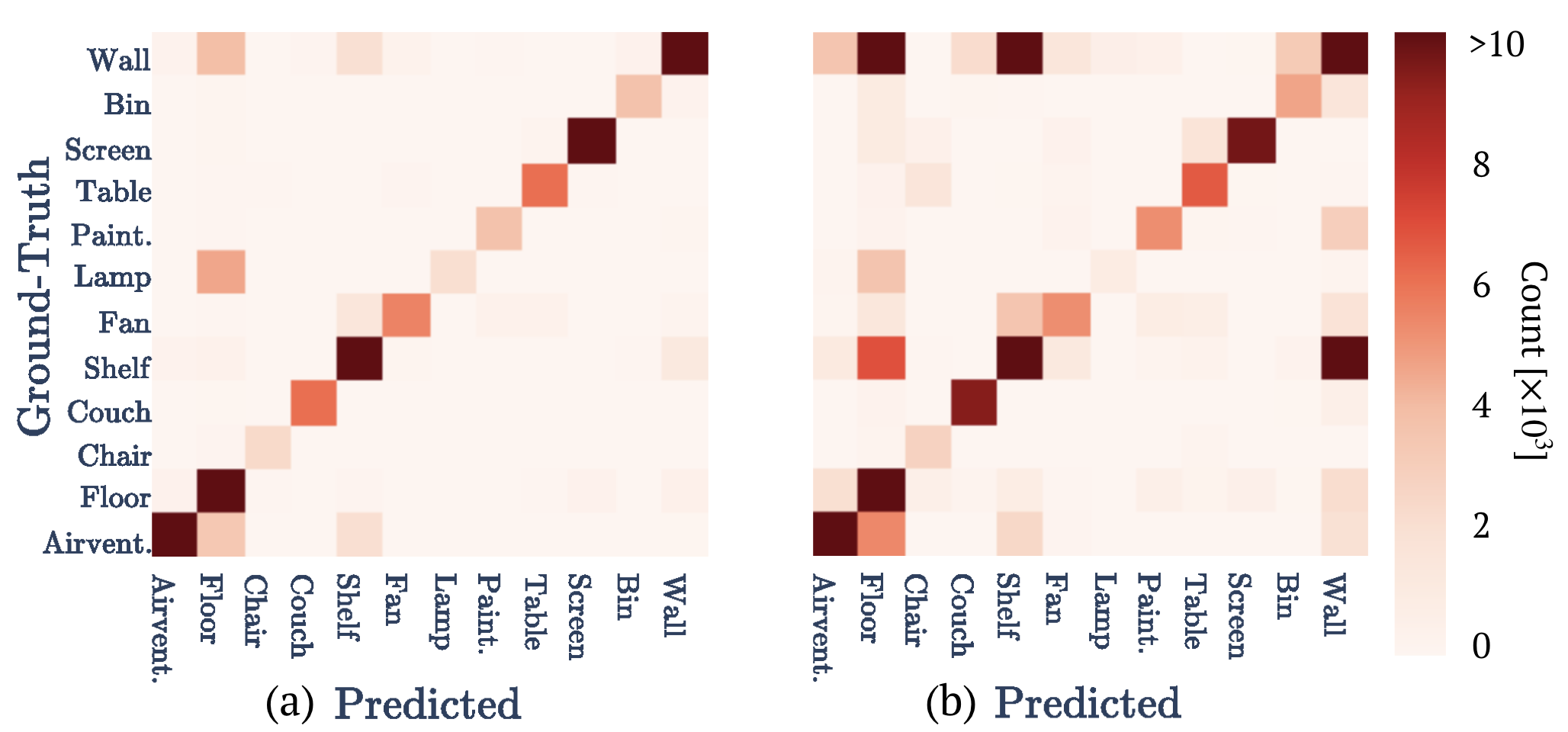}
  \caption{Confusion matrices for \KimeraSemantics using bundled raycasting and (a) ground truth stereo depth or (b) dense stereo~\cite{Hirschmuller08pami}.  Both experiments use ground-truth 2D semantics. 
  Values are saturated to $10^4$ for visualization purposes. \vspace{-2mm}
  \label{fig:confusion_matrix}
  } %
  \vspace{-1mm}
\end{figure}

\KimeraSemantics builds a 3D mesh from the VIO pose estimates, and uses a combination of dense stereo  and 
bundled raycasting. 
We evaluate the impact of each of these components by running three different experiments. 
First, we use \KimeraSemantics with ground-truth (GT) poses and ground-truth depth maps (available in simulation) to assess the initial loss of performance due to bundled raycasting.
Second, we use \KimeraVIO's pose estimates. %
Finally, we use the full \KimeraSemantics pipeline including dense stereo.
To analyze the semantic performance, we calculate the mean Intersection over Union (mIoU)~\cite{Hackel17arxiv-semantic3d}, and the overall portion of correctly labeled points (Acc)~\cite{Wolf15ral}.
We also report the ATE to correlate the results with the drift incurred by \KimeraVIO.
Finally, we evaluate the metric reconstruction 
registering the estimated mesh with the ground truth and computing the RMSE for the points as in Section~\ref{ssec:geometric_performance}.
\Cref{tab:semantic_accuracy} summarizes our findings and shows that bundled raycasting 
results in a small drop in performance both geometrically ($<\!8$cm error on the 3D mesh) as well as semantically (accuracy $>\!94\%$).
Using \KimeraVIO also results in negligible loss in performance since our VIO has a small drift ($<0.2\%$, $4$cm for a $32$m long trajectory).
Certainly, the biggest drop in performance is due to the use of dense stereo.
Dense stereo~\cite{Hirschmuller08pami} %
has difficulties resolving the depth of texture-less regions such as walls, which are frequent in 
 simulated scenes. 
\Cref{fig:confusion_matrix} shows the confusion matrix when running \KimeraSemantics with \KimeraVIO~\Toni{\st{with} and} ground-truth depth (\Cref{fig:confusion_matrix}(a)), compared with using dense stereo (\Cref{fig:confusion_matrix}(b)).
Large  values in the confusion matrix appear between \textit{Wall/Shelf} and \textit{Floor/Wall}.
This is exactly where dense stereo suffers the most; texture-less walls are difficult to reconstruct and are close to shelves and floor, resulting in increased geometric and semantic errors.

\begin{table}[htbp]
  \centering
  \vspace{1mm}
  \caption{Evaluation of \KimeraSemantics. %
  }
  \label{tab:semantic_accuracy}
  \vspace{-1.5mm}
  \begin{tabularx}{\columnwidth}{l c *3{Y}}
    \toprule
     & & \multicolumn{3}{c}{\specialcell[b]{\KimeraSemantics} using:} \\
    \cmidrule(l{2pt}r{2pt}){3-5}
     \multicolumn{2}{c}{Metrics} & \multicolumn{1}{c}{\specialcell[b]{GT Depth \\ GT Poses}} & \multicolumn{1}{c}{\specialcell[b]{GT Depth \\ \KimeraVIO }} & \multicolumn{1}{c}{\specialcell[b]{Dense-Stereo \\ \KimeraVIO }} \\
   \cmidrule(l{2pt}r{2pt}){1-2} \cmidrule(l{2pt}r{2pt}){3-3} \cmidrule(l{2pt}r{2pt}){4-4}  \cmidrule(l{2pt}r{2pt}){5-5}
 \multirow{ 2}{*}{Semantic} & mIoU [\%] & 80.10 & 80.03 & 57.23 \\
    & Acc [\%] & 94.68 & 94.50 & 80.74 \\
    \midrule
 \multirow{ 2}{*}{Geometric} &  ATE [m] & 0.0 & 0.04 & 0.04 \\
   & RMSE [m] & 0.079 & 0.131 & 0.215 \\
    \bottomrule
  \end{tabularx}
  \vspace{-3mm}
\end{table}

\subsection{Timing}
\label{ssec:timing_performance}

\begin{figure}[t!]
  \centering   
  \includegraphics[trim={0cm 0cm 0cm 0cm}, clip, width=1.0\columnwidth]{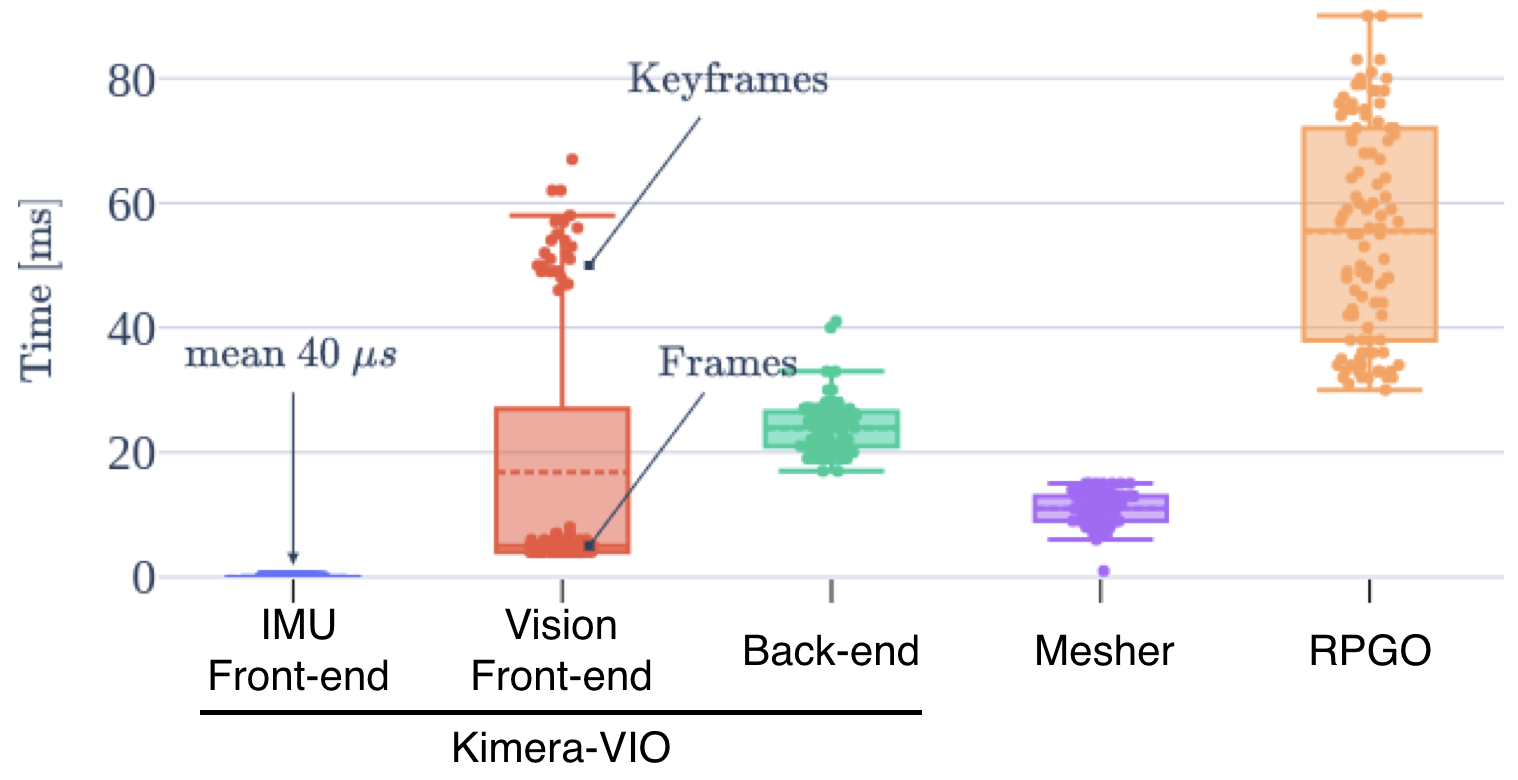} 
  \vspace{-5.5mm} %
  \caption{Runtime breakdown for \KimeraVIO, RPGO, and Mesher.\vspace{-4mm}\label{fig:timing}} 
\end{figure}

\Cref{fig:timing} reports the timing performance of \Kimera's modules.
The IMU front-end requires 
 around $40\mu$s for preintegration, hence can generate state estimates at IMU rate ($>200$Hz \Toni{\st{in our tests}}).
The vision front-end module shows a bi-modal distribution since, for every frame, we just perform feature tracking 
(which takes an average of $4.5$ms), while, 
at keyframe rate, we perform feature detection, stereo matching, and geometric verification,
which\Toni{\st{overall}, combined,} take an average of $45$ms.
\KimeraMesher is capable of generating per-frame 3D meshes in less than $5$ms, while building the multi-frame mesh takes $15$ms on average.
The back-end solves the factor-graph optimization in less than $40$ms.
\KimeraRPGO and \KimeraSemantics run on slower threads since their outputs are not required for time-critical 
actions (\eg control, obstacle avoidance). 
\KimeraRPGO took an average of $55$ms in our experiments on \Euroc, but in general its runtime depends on the size of the pose graph.
Finally, \KimeraSemantics (not reported in figure for  clarity) takes an average of \Toni{\st{$1.28s$}$\globalMeshLatency$} to update the global metric-semantic mesh at each keyframe, fusing a $720 \times 480$ dense depth image, as the one produced by our simulator.
\Toni{\st{Note that Kimera-Semantics can run in $\approx 100$ms without semantic annotations, using the fast raycasting method described in \mbox{\cite{Oleynikova2017iros-voxblox}}.}}

\section{Conclusion}
\label{sec:conclusions}

\Kimera is an open-source C++ library for metric-semantic SLAM. 
It includes state-of-the-art implementations of visual-inertial odometry, robust pose graph optimization, 
  mesh reconstruction, and 3D semantic labeling. 
 It runs in real-time on a CPU and
  provides a suite of continuous integration and 
  benchmarking tools. We hope \Kimera can provide a solid basis for future research on robot perception, 
  and an easy-to-use infrastructure for researchers across communities. %
 %
%

\begin{comment}
\section*{Acknowledgment}

%
We are thankful to Dan Griffith, Ben Smith, Arjun Majumdar, and Zac Ravichandran for kindly sharing the photo-realistic simulator, and to 
 %
 Winter Guerra and Varun Murali for the useful discussions on Unity-based simulations.
 %
%
\end{comment}

%
{\bf Acknowledgments.} We are thankful to Dan Griffith, Ben Smith, Arjun Majumdar, and Zac Ravichandran for kindly sharing the photo-realistic simulator, and to 
 Winter Guerra and Varun Murali for the discussions about Unity. %
\clearpage
\IEEEtriggeratref{63}
\bibliographystyle{IEEEtran}

\end{document}